\documentclass{article}
\usepackage{spconf,amsmath,graphicx}
\usepackage{subfig}
\usepackage{multirow}
\usepackage{hhline}
\usepackage{pgf}
\usepackage{tikz}
\usepackage{tikzscale}
\usepackage{url}
\usetikzlibrary{arrows,automata, positioning, fit, backgrounds}

\newcommand{\specialcell}[2][c] 
{%
  \begin{tabular}[#1]{@{}l@{}}#2\end{tabular}
}

\usepackage{caption}

\usepackage{fancyhdr}

\title{End-to-end Person Search Sequentially Trained on Aggregated Dataset}
\name{Angelique Loesch$^{\star, \dagger}$ \qquad Jaonary Rabarisoa$^{\star, \dagger}$ \qquad Romaric Audigier$^{\star, \dagger}$
\thanks{This research is supported by Conseil regional d'Ile-de-France and BpiFrance through the COOPOL and ETS projects.}
}

\address{$^{\star}$ Université Paris-Saclay, CEA, List, F-91120, Palaiseau, France\\
    $^{\dagger}$Vision Lab, ThereSIS, Thales SIX GTS, Campus Polytechnique, Palaiseau, France\\
    \{angelique.loesch, jaonary.rabarisoa, romaric.audigier\}@cea.fr\\}
\begin{document}
%
\maketitle
\begin{abstract}
In video surveillance applications, person search is a challenging task consisting in detecting people and extracting features from their silhouette for re-identification (re-ID) purpose.
We propose a new end-to-end model that jointly computes detection and feature extraction steps through a single deep Convolutional Neural Network architecture.
Sharing feature maps between the two tasks for jointly describing people commonalities and specificities allows faster runtime, which is valuable in real-world applications.
In addition to reaching state-of-the-art accuracy, this multi-task model can be sequentially trained task-by-task, which results in a broader acceptance of input dataset types.
Indeed, we show that aggregating more pedestrian detection datasets without costly identity annotations
makes the shared feature maps more generic, and improves re-ID precision.
Moreover, these boosted shared feature maps result in re-ID features more robust to a cross-dataset scenario.
\end{abstract}
\section{Introduction}
\label{sec:intro}

\emph{Person re-identification (re-ID)}~\cite{Liu2017a,Liao2015,Li2014} is an essential task in video surveillance 
that has gained much attention over the last decade in academic research.
It consists in recognizing a person represented by a query (``probe'') image snippet in a set (``gallery'') of image snippets of people.
However, in real use-case scenarios such as perpetrator search, cross-camera person tracking or person activity analysis, image snippets around the people are not available. They have to be extracted from the full scene images of interest.
Thus, re-ID results also depend on the quality of a detector that localizes all the people in the scene. 
\emph{Person search} is the problem considering both detection and re-ID tasks in a unique framework or system.
Person search approaches can be divided into two categories: disjoint (sequential) methods and joint (end-to-end) methods.

On the one hand, \emph{disjoint methods} sequentially detect people then extract re-ID features. 
Zheng et al.~\cite{Zheng2017} show that the results of the identity matching task is directly correlated to the alignment quality of the detected bounding boxes. 
Thus, several approaches~\cite{Schumann2017,Lan2018,Chen2018b} separately train detection and re-ID modules before running them in a pipeline. 
Disjoint methods can benefit from any improvement of state-of-the-art in pedestrian detection (possibly given by a multi-class object detector) and in snippet re-ID~\cite{Li2017,Chen2018b,Yu2017,Lan2018}.
However, in order to fulfill operational requirements, a tradeoff must be made between accuracy and runtime of the selected modules: indeed, the best detectors and re-ID feature extractors have in general higher computational cost.

On the other hand, \emph{joint methods} for person search propose single end-to-end Convolutional Neural Network (CNN) architectures where both detection and re-ID tasks are jointly handled from a full scene image.
%
%
\begin{figure*}[!ht]
  \centering
  \includegraphics[width=1\linewidth]{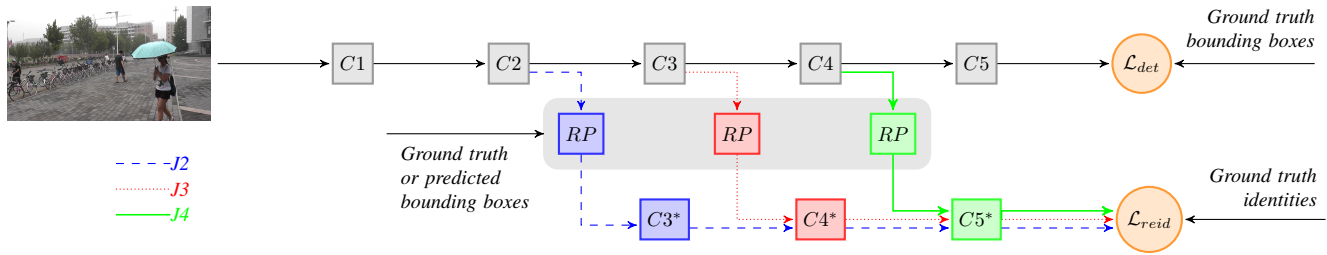}
  \caption{
Proposed multi-task architecture based on ResNet-50 \cite{He2016}. 
$C1$ is the first convolution layer. 
$C2$ to $C4$ are the bottleneck blocks.
Blocks $C1$ to $C2$ (resp. $C1$ to $C3$, or $C1$ to $C4$) are shared between detection and re-ID branches for model variant named $J2$ in dashed blue (resp. $J3$ in dotted red, or $J4$ in solid green).
The re-ID branch begins with a ROI-pooling ($RP$) layer and remaining replicated bottleneck blocks $C3^*$ (resp. $C4^*$, or $C5^*$) to $C5^*$. 
The example image is from PRW \cite{Zheng2017}.}
  \label{fig:joint_model}
\end{figure*}
%
%
An original approach~\cite{Liu2017e} implements recursive localization and search refinement to more accurately locate the target person in the scene. 
However, the use of convolutional LSTM \cite{Xingjian2015} has scalability issues and is thus not easily applicable to real-case scenarios.
Most joint methods are based on the two-stage detector Faster-RCNN~\cite{Ren2017}.
Their architectures are composed of a shared convolutional layer backbone whose resulting feature maps are shared by two distinct parts: a pedestrian proposal net as the Region Proposal Network, and an identification net to classify among identities \cite{Xiao2016a,Xiao2017,Xiao2019,Chen2018a,He2018,Liu2018,Shi2018}. 
Both parts are jointly optimized on train datasets of full images annotated with bounding boxes and identities (e.g. CUHK-SYSU \cite{Xiao2016b} or PRW \cite{Zheng2017} datasets).
The re-ID task is formulated as a classification problem. 
In order to work around conventional softmax loss drawback 
and to exploit the unlabeled identities with no specific class {ID}s, 
Online Instance Matching (OIM) loss~\cite{Xiao2017} or Instance Enhancing Loss (IEL)~\cite{Shi2018}  are used,
enabling faster and better convergence. 
Other methods propose to fuse these losses with a center loss~\cite{Xiao2019,Liu2018}, or Hard Example Priority based softmax loss (HEP)~\cite{He2018}.
Joint methods can obtain equivalent performance as disjoint ones when using architectures of comparable complexity~\cite{Xiao2017}
while the use of a shared backbone by joint methods significantly decreases runtime.
Nonetheless, joint methods need train datasets with both annotation types (people bounding boxes and IDs) which are fewer than pedestrian and snippet re-ID datasets.  On the contrary, disjoint methods can equally use datasets with one or both annotation types. This broader dataset acceptance is an advantage if more data is needed for greater genericity and robustness against dataset biases.
%
%

The contributions of this article to the person search problem are as follows:
(1) We first propose a \emph{new end-to-end CNN architecture based on a single-shot detector} (SSD) architecture~\cite{Liu2016a}.
Unlike state-of-the-art methods, we address the re-ID task through the use of \emph{triplet loss for metric learning} which has shown better results than classification loss~\cite{Hermans2017}.
The proposed architecture is competitive with state-of-the-art methods on PRW and CUHK-SYSU datasets.
%
(2) Besides, as runtime is important in real-case applications, a study is carried out to assess the \emph{tradeoff between runtime and performance w.r.t. the shared backbone size}.
%
(3) Furthermore, \emph{sequentially training} the two joint subnets of our model allows the \emph{aggregation of more train datasets} for people detection. We show that training the detection task with more data leads to better performance in re-ID. The shared backbone produces feature maps that seem to better describe people commonalities and specificities.
%
(4) Finally, first results show that feature maps learned from such aggregated detection datasets also lead to better re-ID performance when applied to \emph{cross-dataset} scenarios, i.e, when the target dataset 
is not seen during training.
Cross-dataset scenarios are of utmost importance for real use-cases.
Indeed, no end-user can afford to annotate identities on operational environment because it is too fastidious and time-consuming. 
%

\vspace{-0.3cm}
\section{Proposed Method}
\label{sec:method}

We propose a \emph{multi-task} architecture to jointly solve detection and re-ID tasks.
We build this architecture with the following guidelines:
(1) Use \emph{SSD} and keep the performance of the detection task as high as possible. Single shot detectors are computationally efficient and can be very accurate when fine-tuned on targeted domain and task (i.e. specialized for the single class 'people')
(2) Implement multi-task as \emph{hard parameter sharing}~\cite{Ruder17a} to reduce forward complexity. 
(3) Use \emph{triplet loss} to solve the re-ID task as it is an effective way to learn representation. 
(4) Make it possible to \emph{use different dataset types}. Existing datasets for joint detection and re-ID are relatively small. Training the detector on these datasets alone generally results in poor detection performance in cross-dataset. With a \emph{two-step training}, we can use all available detection data along with joint detection and re-ID annotated data.
\vspace{-0.3cm}
\paragraph*{Architecture}
Our architecture is based on SSD~\cite{Liu2016a} on which we add a branch for the re-ID task.
The detection and the re-ID subnets share common backbone layers. 
In this study, we use the Retina\-Net \cite{Tsung2017} architecture with a ResNet-50 \cite{He2016} as feature extractor. 
The first convolution and ResNet blocks are shared between both branches. 
We keep the same architecture as RetinaNet for the rest of the detection subnet. 
On the other hand, the re-ID subnet is composed of a ROI-pooling and the replication of remaining ResNet blocks. During the training phase, the network accepts two different types of data: \textit{image} with \textit{bounding boxes} to train the detection branch or \textit{image} with \textit{identified bounding boxes} to train the re-ID branch. This way, we can use all available detection datasets and joint person search datasets to train the network in two different steps. 
The joint architecture with 3 different layer sharings between branches are depicted in Fig.~\ref{fig:joint_model}. 
These variants are denoted by $J2$, $J3$ and $J4$ in the following.
The number of the shared layers will impact the architecture complexity and runtime. 
The more layers shared by the two branches the faster it will be at runtime.
\vspace{-0.3cm}
\paragraph*{Training objective}
To train the detection branch we follow the RetinaNet approach and use the same objective. The training loss $\mathcal{L}_{det}$ is the sum of the \textit{focal} loss and the standard \textit{smooth $L^1$} loss. We refer the reader to \cite{Tsung2017} for more details.
For the re-ID branch we use the triplet loss as objective function $\mathcal{L}_{reid}$. Precisely, we take the batch-hard formulation proposed by \cite{Hermans2017}. However, a good initialization is necessary in order to avoid the trivial null function solution. 
Thus, we pre-train the network with the semi-hard triplet \cite{Song2017} loss.
The global training objective is: 
$\mathcal{L} = \mathcal{L}_{det} + \mathcal{L}_{reid}$.
\vspace{-0.3cm}
\paragraph*{Two-step training}
Classically, to train our multi-task network we have to minimize $\mathcal{L}$ directly. 
Nevertheless, to keep detection at its highest precision we set the input image size at 640x640. 
This reduces the number of images per batch that we can feed in the network and makes difficult the minimization of $\mathcal{L}_{reid}$.
To overcome this problem, we follow a two-step training strategy. 
First, we train the detection branch until convergence using the detection data only. 
Then, we pre-compute the feature maps shared between the detection and re-ID branches and pool the features of all ground-truth bounding boxes of the re-ID dataset. 
Finally, we train the re-ID branch using these pooled features as input.
This two-step approach reduces the memory footprint of the re-ID branch during the training phase and enables increased batch sizes.
This ensures that the algorithm finds more informative triplets while minimizing $\mathcal{L}_{reid}$. 
Notice that gradients from $\mathcal{L}_{reid}$ are not back-propagated to the shared layers
in order to keep detection performance. 
\vspace{-0.05cm}
\section{Experiments and Results}
\label{sec:Experiments}
\subsection{Experiment settings}
\label{ssec:settings}
\paragraph*{Datasets}
Either CUHK-SYSU~\cite{Xiao2016b} or PRW~\cite{Zheng2017} dataset is used to train and evaluate our model.
CUHK-SYSU (resp. PRW) contains 18,184 movies and street surveillance images (resp. 11,816 outdoor images) with 99,809 (resp. 34,304) annotated person bounding boxes of 8,432 (resp. 932) unique identities, being about 12 (resp. 37) boxes per ID.
It is divided in a train set of 11,206 (resp. 5,704) images with 5,532 (resp. 483) identities, and a test set of 6,978 (resp. 6,112) gallery frames and 2,900 (resp. 449) query people. 
To improve performances, two pedestrian datasets can be added during the first-stage of training (detection part): MOT17Det \cite{Mot17}
with 5,316 train images and 112,297 annotated bounding boxes, and Wider Pedestrian dataset \cite{Wider18}
 with 11,500 train images and 46,513 bounding boxes.
\vspace{-0.28cm}
\paragraph*{Implementation Details}
Our architecture is based on RetinaNet with Resnet-50 feature extractor on which we add the re-ID branch. We set the image size to 640x640. 
The detector is fine-tuned from weights pre-trained on MSCOCO dataset. We use a mini-batch of size 10 and Stochastic Gradients Descent with momentum $0.9$. Learning rate follows a linear-cosine decay scheme with warm up and a base value at $10^{-3}$. Number of training epochs is $90$.
To train the re-ID branch we sample mini-batches of pooled features with $32$ different identities and $4$ shots per identity. We use ADAM optimizer and learning rate with a linear-cosine decay scheme starting at $10^{-4}$. Re-ID branch is trained with semi-hard triplet loss ($350$ epochs) then with batch-hard triplet ($350$ epochs).
\vspace{-0.3cm}
\paragraph*{Evaluation Protocols}
The proposed method is evaluated following CUHK-SYSU \cite{Xiao2016a} and PRW \cite{Zheng2017} protocols. 
Both of them consider a predicted bounding box as positive if its overlap with the ground truth box is greater than 0.5, and use mean Average Precision (mAP) and rank-1 matching rate (Rank-1) metrics. 
Yet, protocols slightly differ: 
CUHK-SYSU considers galleries of increasing size. We report results for 100-image (the reference in literature) and 4,000-image (the largest one) galleries.
As for PRW protocol, it keeps a fixed gallery size of 6,112 images, but computes mAP and Rank-1 
w.r.t. the number of detected boxes per image (by increasing detector recall).
We report best results for each method when varying this number of boxes per image.

\vspace{-0.25cm}
\subsection{Results}
\label{ssec:intra_evaluation}
\vspace{-0.1cm}
\paragraph*{Influence of shared backbone size on re-ID performance and computation time}
Table~\ref{tab:comp_time} shows mean computation time  w.r.t. the shared backbone size.
Times were measured on a Titan X GPU with different batch sizes and number of people in the image.
$Disj.$ is a disjoint method with comparable architectures (a Retina\-Net \cite{Tsung2017} followed by a ResNet-50 feature extractor).
With 5 people (resp. 20 people) per image, even with a short shared backbone as in $J2$, our model can be 1.4 to 1.7 (resp. 1.9 to 2.5) faster than the related disjoint method, according to the number of images per batch.
With a longer shared backbone as in $J4$, the gain is more significant, our method being 1.5 to 2.0  (resp. 2.5 to 3.4) faster.
Joint models are closer to fulfill the real-world requirements than the disjoint one, especially when image/people batches are used.
However re-ID precision depends on the difficulty level of the test dataset. 
Table \ref{tab:prw_sysu_sota_error} (top) shows mAP and Rank-1 results of the proposed method on both datasets w.r.t. the shared backbone size.
On CUHK-SYSU, the 3 variants have equivalent results
even with one single ResNet block specialized for re-ID ($J4$ variant).
$J3$ shows the best bias/variance trade-off. 
But, on the more difficult PRW dataset, 
one block is not enough to solve the re-ID task and $J4$ is far less accurate than $J2$ and $J3$.
Thus $J3$, dealing great with both datasets, seems to achieve the best tradeoff accuracy/runtime.
%
%
\begin{table}[!t]
\centering
\begin{minipage}[!b]{0.9\linewidth}
\subfloat[b][
\label{tab:prw_sysu_sota_error}]{
\resizebox{\linewidth}{!}{
\begin{tabular}{l|*{3}{c}*{1}{c|}}
 \cline{2-5}
\multicolumn{1}{c|}{}&\multicolumn{2}{c}{\multirow{2}{*}{PRW}} & \multicolumn{2}{||c|}{CUHK-SYSU}\\
\cline{4-5}
& & & \multicolumn{2}{||c|}{gallery size 100 / 4000}\\
\cline{2-5}
\multicolumn{1}{c|}{} & \small{mAP (\%)} & \small{Rank-1 (\%)} &\multicolumn{1}{||c}{ \small{mAP (\%)}} & \small{Rank-1 (\%) } \\\hline
\multicolumn{1}{|c|}{$Disj.^\ddag$} & 13.3 & 32.3&\multicolumn{1}{||c}{72.1 / 50.1} & 74.1 / 53.3\\\hline
\multicolumn{1}{|c|}{$J2$ (ours)$^\ddag$}& \textbf{25.2} & 47.0&\multicolumn{1}{||c}{76.4 /  49.2} & 76.7 / 51.3\\\hline
\multicolumn{1}{|c|}{$J3$ (ours)$^\ddag$} &  22.5 & 45.1&\multicolumn{1}{||c}{79.4 / 55.8}& 80.5 / \textbf{58.9} \\\hline
\multicolumn{1}{|c|}{$J4$ (ours)$^\ddag$} & 12.3 & 27.3& \multicolumn{1}{||c}{76.7 / 53.3} & 77.8 / 56.0
\\\hhline{|=|==|==|} 
\multicolumn{1}{|c|}{\small{Xiao2016 \cite{Xiao2016a}}} & - & - &\multicolumn{1}{||c}{55.7 / -} & 62.7 / 42.5\\\hline
\multicolumn{1}{|c|}{\small{JDI+OIM \cite{Xiao2017}$^\ddag$}} &  21.3 & 49.9 &\multicolumn{1}{||c}{75.5 / 51.0} &  78.7 / -\\\hline
\multicolumn{1}{|c|}{\small{IAN \cite{Xiao2019}$^{*}$}} &  23.0 & 61.8 &\multicolumn{1}{||c}{77.2 / 55.0} & 80.7 / -\\\hline
\multicolumn{1}{|c|}{\small{Chen 2018 \cite{Chen2018a}}} & - &- &\multicolumn{1}{||c}{78.8 / -} & 80.9 / -\\\hline
\multicolumn{1}{|c|}{\small{I-Net \cite{He2018}}} & - &- &\multicolumn{1}{||c}{79.5 / 53.5} & \textbf{81.5} / -\\\hline
\multicolumn{1}{|c|}{\small{Liu2018 \cite{Liu2018}$^{*}$}} &  21.0 & 63.1 &\multicolumn{1}{||c}{\textbf{79.8} / -} &  79.9 / -\\\hline
\multicolumn{1}{|c|}{\small{JDI+IEL \cite{Shi2018}$^{*}$}} &  24.3 & \textbf{69.5} &\multicolumn{1}{||c}{ 79.4 / \textbf{58.0}} &79.7 / -\\\hline
\multicolumn{1}{|c|}{\small{NPSM \cite{Liu2017e}$^\ddag$}} & 24.2 &53.1 &\multicolumn{1}{||c}{77.9 / 54.0} & 81.2 / -
\\\hline
\multicolumn{5}{l}{
Highest score reported for PRW protocol at 
} \\
\multicolumn{5}{l}{
$^{*}$: 3 bounding boxes / image; $^\ddag$: 5 bounding boxes / image.
}
\end{tabular}
}
}
\end{minipage}
\vspace{0.35cm}
\\
\hspace{-0.1cm}
\begin{minipage}[!b]{0.5\linewidth}
\subfloat[b][\label{tab:comp_time}]{
\resizebox{\linewidth}{!}{
\begin{tabular}{|l|*{1}{c||}*{2}{c|}}
 \cline{2-4}
\multicolumn{1}{c|}{} & \multirow{2}{*}{$\frac{\# im.}{batch}$}   & \multicolumn{2}{c|}{computation time (ms)} \\\cline{3-4}
\multicolumn{1}{c|}{}& &  \small{5 p. / im.} & \small{20 p. / im.}\\\cline{1-4}
\multirow{3}{*}{\small{$Disj.$}} & 1  &17.0 &7.4\\\cline{2-4}
&4 &13.6&6.6 \\\cline{2-4}
&8 &12.9&6.4 \\\hhline{|=|=|=|=|}
\multirow{3}{*}{$J2$}& 1  &12.3  &3.9 \\\cline{2-4}
&4 &8.3 & 2.7\\\cline{2-4}
&8 &7.4 &2.5 \\\hhline{|=|=|=|=|}
\multirow{3}{*}{$J3$}& 1  & 12.2 & 3.4\\\cline{2-4}
&4 &7.8 &2.4\\\cline{2-4}
&8 & 7.2&2.2 \\\hhline{|=|=|=|=|}
\multirow{3}{*}{$J4$}& 1  & \textbf{11.1} &\textbf{2.9} \\\cline{2-4}
&4 &\textbf{7.1} &\textbf{1.9} \\\cline{2-4}
&8 &\textbf{6.5} & \textbf{1.9}\\\cline{1-4}
\end{tabular}
}
}
\end{minipage}
\hspace{-0.2cm}
\begin{minipage}[!b]{0.5\linewidth}
\vspace{1.5cm}
\subfloat[][\label{tab:sysu_transfer_error}]{
\resizebox{\linewidth}{!}{
\begin{tabular}{|l|*{2}{c|}}
 \cline{2-3}
\multicolumn{1}{c|}{} & \multicolumn{2}{c|}{gallery size 100 / 4000}\\\cline{2-3}
\multicolumn{1}{c|}{} & \small{mAP (\%)} & \small{Rank-1 (\%)}\\\hline
$J2_{\text{p}}$ &  31.5 / 14.9 & 33.4 / 16.1\\\hline
$J2_{\text{m-w-p}}$ & \textbf{54.4 / 29.4} & \textbf{55.4 / 31.9}\\\hhline{|=|=|=|}
$J3_{\text{p}}$ & 29.8 / 13.9& 31.7 / 15.8\\\hline
$J3_{\text{m-w-p}}$ & 54.6 / 28.1& 56.1 / 29.7\\\hhline{|=|=|=|}
$J4_{\text{p}}$ &  22.8 / 8.8&23.1 / 9.5\\\hline
$J4_{\text{m-w-p}}$ & 52.5 / 27.8& 53.3 / 28.8\\\hline
\end{tabular}
}
}
\end{minipage}
\\
\subfloat[][\label{tab:sysu_det_data_error}]{
\resizebox{0.9\linewidth}{!}{
\begin{tabular}{|l|*{1}{c}*{1}{c||}*{2}{c}}
 \cline{2-5}
\multicolumn{1}{c|}{} & \multicolumn{4}{c|}{gallery size 100 / 4000} \\ \cline{2-5}
\multicolumn{1}{c|}{} & \small{mAP (\%)} & \small{Rank-1 (\%)}& \small{mAP GT (\%)} &\multicolumn{1}{c|}{ \small{Rank-1 GT (\%)}}  \\\hline
\specialcell{$J2_{\text{c}}$} & 71.4 / 43.6& 71.6 / 45.5 & 78.6 / 50.3&\multicolumn{1}{c|}{78.0 / 52.3}\\\hline
\specialcell{$J2$} & 76.4 / 49.2&76.7 / 51.3& 81.9 / 54.9& \multicolumn{1}{c|}{81.0 / 56.5}\\\hhline{|=|==||==|}
\specialcell{$J3_{\text{c}}$} & 75.5 / 48.1 & 76.4 / 50.3&81.2 / 54.2& \multicolumn{1}{c|}{80.9 / 56.5}\\\hline
\specialcell{$J3$} & \textbf{79.4} / \textbf{55.8}& \textbf{80.5} / \textbf{58.9}&\textbf{84.4} / \textbf{60.9}& \multicolumn{1}{c|}{\textbf{84.0} / \textbf{63.1}}\\\hhline{|=|==||==|}
\specialcell{$J4_{\text{c}}$} & 62.9 / 33.3 &62.3 / 33.8&68.5 / 37.1&\multicolumn{1}{c|}{ 67.1 / 37.2} \\\hline
\specialcell{$J4$} & 76.7 / 53.3& 77.8 / 56.0&81.6 / 57.1&\multicolumn{1}{c|}{ 81.3 / 58.8}\\\hline
\end{tabular}
}
}
\vspace{0.1cm}
\caption{
\textbf{(a)} Mean average precision (mAP) (\%) and matching rate at rank-1 (Rank-1) (\%) for  \textbf{(top)} our joint models $J\text{x}$, a comparable disjoint baseline $Disj.$ and \textbf{(bottom)} state-of-the-art joint methods on PRW and CUHK-SYSU.\\
\textbf{(b)} Mean computation time (ms) to detect a person and extract his/her re-ID feature w.r.t. shared backbone size, (full image) batch size and number of people in the image (snippet image batch).\\
\textbf{(c)} mAP and Rank-1 on CUHK-SYSU cross-dataset for our joint models $J\text{x}_{\text{p}}$ trained on PRW only, or for $J\text{x}_{\text{m-w-p}}$ boosted by pedestrian dataset aggregation. NB: CUHK-SYSU was not used during training.\\
\textbf{(d)} 
\textbf{(left)} mAP and Rank-1 on CUHK-SYSU for our joint models 
$J\text{x}_{\text{c}}$ trained on CUHK-SYSU only, or for $J\text{x}$ boosted by pedestrian dataset aggregation. 
\textbf{(right)} Same with injection of ground truth (GT) boxes instead of predicted boxes. }
\vspace{-0.65cm}
\label{fig:tables}
\end{table}

\vspace{-0.3cm}
\paragraph*{Boosting shared feature map efficiency for intra-dataset re-ID}
%
%
%
%
%
Table \ref{tab:sysu_det_data_error} (left) shows mAP and Rank-1 improvement on CUHK-SYSU dataset, when using our sequential training on aggregated pedestrian dataset.
$J2_\text{c}$, $J3_\text{c}$, $J4_\text{c}$ denote the variants which are trained on CUHK-SYSU only.
$J2$, $J3$, $J4$ denote the variants which are trained on an aggregated detection dataset (MOT17Det, Wider, CUHK-SYSU and PRW) for the detection branch, then on CUHK-SYSU only for the re-ID branch. 
When shared feature maps are boosted by aggregated pedestrian dataset, 
$J2$, $J3$ and $J4$ mAP rise 5\%, 4\% and 14\% with a 100-image gallery. The influence of aggregated dataset is clearer for longer backbone (i.e. $J4$).
This improvement is even clearer on a more difficult gallery (4000 images: +20\% for $J4$ compared to $J4_\text{c}$).
To be sure
the dataset aggregation enhances the re-ID performance through the learned feature maps and not only through the bounding box precision, we evaluate the same models with the injection of ground truth (GT) boxes (cf. Table \ref{tab:sysu_det_data_error} right).
Again, aggregation makes feature maps more efficient (+13\% or +20\% mAP improvement between $J4$ and $J4_\text{c}$ for both galleries).
\vspace{-0.3cm}
\paragraph*{Boosting shared feature map genericity for cross-dataset re-ID}
Similar experiments are performed on a cross-dataset scenario to highlight the interest for shared feature map boosting:
$J2_\text{p}$, $J3_\text{p}$, $J4_\text{p}$ are trained on PRW only, whereas 
$J2_{\text{m-w-p}}$, $J3_{\text{m-w-p}}$, $J4_{\text{m-w-p}}$ are trained on aggregated pedestrian dataset (MOT17Det, Wider and PRW) for detection branch, and on PRW only, for re-ID branch (same settings as in Section \ref{ssec:settings}).
As expected, aggregated dataset brings more genericity to the detector: Average Precision (AP) for $J\text{x}_\text{p}$ is 81\% on PRW (intra-dataset) and 30\% on CUHK-SYSU (cross-dataset) whereas AP for $J\text{x}_{\text{m-w-p}}$ is 79\% on PRW (intra-dataset) and 60\% on CUHK-SYSU (cross-dataset).
More impressively, shared feature maps boosted by pedestrian datasets also bring robustness to cross-dataset re-ID (cf. Table~\ref{tab:sysu_transfer_error}): 
%
%
on CUHK-SYSU cross-dataset, mAP and Rank-1 increase from 22 to 30\% (resp. 14\% to 19\%) on a 100-image (resp. 4000-image) gallery for all 3 boosted variants. 
The feature map genericity clearly helps cope with cross-dataset re-ID.
Thus, aggregating pedestrian datasets during sequential training turns out to be a not costly yet efficient way to increase re-ID performance on real-case cross-domain scenarios, when neither data nor identities are available from the target use case.

\vspace{-0.2cm}
\paragraph*{Comparison with person search state-of-the-art}
\label{ssec:ps_evaluation}
Table \ref{tab:prw_sysu_sota_error} compares the proposed method (at the top) with joint person search state-of-the-art approaches (at the bottom) on PRW and CUHK-SYSU datasets. 
Overall, our method compares well with state-of-the-art:
On PRW, although Rank-1 for $J2$ is not so high, mAP slightly outperforms state-of-the-art (+0.9\% over JDI+IEL \cite{Shi2018}).
This means that even if the first match is not always correct,
an overall good recall is obtained for all shots of the query person.
On CUHK-SYSU, $J3$ reaches the third best mAP (at 0.4\% below top-1~\cite{Liu2018}) on the 100-image gallery, while Rank-1 is similar to state-of-the-art approaches
(at 1\% below best rank-1~\cite{He2018}).
On the 4000-image gallery, our method $J3$ obtains the second best mAP results (at 2.2\% below JDI+IEL~\cite{Shi2018}). 
\vspace{-0.2cm}
\section{Conclusion}
\label{sec:conclusion}
In this article, new end-to-end person search networks are presented, based on SSD architecture for the detection task, and triplet loss to solve the re-ID task by metric learning.
A study is carried out on 3 different variants to assess the precision/runtime tradeoff w.r.t. the shared backbone size.
Our method reaches competitive 
re-ID results on CUHK-SYSU and PRW datasets, 
compared to other joint state-of-the-art approaches.
Moreover, we show that aggregating pedestrian datasets during the sequential training leads to significant improvement in intra and cross-dataset scenarios. 
Pedestrian datasets being more widely spread than person search datasets, 
the proposed methodology for boosting shared feature maps turns out to be
very useful for real-world applications.
The exploitation of classic data augmentation (e.g. flip, crop, etc.) techniques could also improve these results. 



\bibliographystyle{IEEEbib}
\bibliography{version_2}

\end{document}